\documentclass[twoside]{article}
\usepackage[preprint]{aistats2026}
\usepackage[round]{natbib}

\usepackage[T1]{fontenc}
\usepackage[utf8]{inputenc}
\usepackage[table]{xcolor}
\usepackage{subcaption}
\usepackage{graphicx}
\usepackage{dsfont} 
\usepackage{amsmath, amssymb, amsthm}
\usepackage{algorithm}
\usepackage{algpseudocode}
\usepackage{tikz}
\usepackage{tikz-cd}
\usetikzlibrary{cd,arrows,calc,shapes,positioning}
\usepackage{tabularx}
\usepackage{booktabs}
\usepackage{hyperref}
\hypersetup{
  unicode=true,           %
  bookmarksopen=true,
  bookmarksnumbered=true
}

\usepackage{enumitem}
\makeatletter
\providecommand*{\theHalgorithm}{\arabic{algorithm}}
\providecommand*{\theHALG@line}{\theHalgorithm.\arabic{ALG@line}}
\makeatother
\usepackage[capitalise]{cleveref}

\crefname{figure}{Figure}{Figures}
\Crefname{figure}{Figure}{Figures}

\newcommand{\G}{\ensuremath G}
\renewcommand{\H}{\ensuremath H}
\newcommand{\aid}{d_{\mathcal{I}}}
\newcommand{\aidbar}{\bar{d}_{\mathcal{I}}}
\newcommand{\aidsym}{d^{\mathrm{sym}}_{\mathcal{I}}}
\newcommand{\aidsymbar}{\bar{d}^{\mathrm{sym}}_{\mathcal{I}}}
\newcommand{\given}{\mid}
\newcommand{\cC}{\mathcal{C}}
\newcommand{\cD}{\mathcal{D}}
\newcommand{\cI}{\mathcal{I}}
\newcommand{\cV}{\mathcal{V}}

\newcommand{\bL}{\mathbf{L}}
\newcommand{\bS}{\mathbf{S}}
\newcommand{\bV}{\mathbf{V}}

\newcommand{\bW}{\mathbf{W}}
\newcommand{\bw}{\mathbf{w}}
\newcommand{\x}{\mathbf{x}}

\newcommand{\Y}{\mathbf{Y}}
\newcommand{\y}{\mathbf{y}}
\newcommand{\bZ}{\mathbf{Z}}
\newcommand{\D}{\mathbf{D}}

\newcommand{\bR}{\mathbb{R}}

\newcommand{\intr}{\mathrm{In}}
\newcommand{\ch}{\ensuremath \mathrm{ch}}
\newcommand{\pa}{\ensuremath \mathrm{pa}}
\newcommand{\de}{\ensuremath \mathrm{de}}
\newcommand{\nd}{\ensuremath \mathrm{nd}}
\newcommand{\an}{\ensuremath \mathrm{an}}
\newcommand{\mb}{\ensuremath \mathrm{mb}}

\newcommand{\dis}{\ensuremath \mathrm{dis}}

\newcommand{\ndG}[1]{\ensuremath \mathrm{nd}^{\G}_{#1}}

\tikzstyle{obs} = [circle,fill=white,draw=black,inner sep=1pt,minimum size=25pt,node distance=0.75cm,thick]
\tikzstyle{latent} = [obs,dotted]
\tikzstyle{fixed} = [obs,rectangle,minimum size=22pt]
\tikzstyle{small} = [inner sep=0pt,minimum size=20pt]

\newcommand{\edge}[3][]{%
  \foreach \x in {#2} {%
    \foreach \y in {#3} {%
      \path (\x) edge [->, >={triangle 45}, thick, #1] (\y) ;%
    };
  };
}

\newtheorem{definition}{Definition}

\newtheorem{theorem}{Theorem}

\begin{document}
\twocolumn[

\aistatstitle{Graph Distance Based on Cause-Effect Estimands with Latents}

\aistatsauthor{Zhufeng Li\textsuperscript{1,2,3} \And Niki Kilbertus\textsuperscript{1,2,3}}
\runningauthor{Zhufeng Li, Niki Kilbertus}
\aistatsaddress{
  \textsuperscript{1} Helmholtz Munich\\
  \textsuperscript{2} Technical University of Munich\\
  \textsuperscript{3} Munich Center for Machine Learning
}
]

\begin{abstract}
  Causal discovery aims to recover graphs that represent causal relations among given variables from observations, and new methods are constantly being proposed. Increasingly, the community raises questions about how much progress is made, because properly evaluating discovered graphs remains notoriously difficult, particularly under latent confounding. We propose a graph distance measure for acyclic directed mixed graphs (ADMGs) based on the downstream task of cause-effect estimation under unobserved confounding. Our approach uses identification via fixing and a symbolic verifier to quantify how graph differences distort cause-effect estimands for different treatment-outcome pairs. We analyze the behavior of the measure under different graph perturbations and compare it against existing distance metrics.
\end{abstract}

\section{Introduction}\label{sec:intro}

\subsection{Motivation}

Uncovering causal relationships from observational data, often called causal discovery or structure learning, is a longstanding goal within causal inference.
In particular, causal discovery is a well-defined specific task within the structural causal model (SCM) framework popularized by Pearl \citep{Pearl2000causality}.
There, the assumed data generating process, the SCM, entails an observational distribution as well as a directed acyclic graph (DAG) over the observed variables.
The task is then (i) to decide whether the graph, or which parts of it, can be uniquely identified given the observational distribution (this is called \emph{identifiability}), and (ii) devise algorithms to actually estimate a graph from finite observations.
Both parts have received considerable attention from the scientific community \citep{heinze2018causal,glymour2019review,zanga2022survey,vowels2022d}.
While there has been no shortage of newly proposed methods---leveraging major advances such as automatic differentiation for differentiable causal discovery (e.g., \citealp{zheng2018dags}), variational inference for Bayesian approaches (e.g., \citealp{annadani2021variational}), amortized causal discovery via large scale pre-training (e.g., \citealp{lorch2022amortized}), or a rapidly growing body of work supporting causal discovery process with large language models---a common dictum in the community is that ``causal discovery is still awaiting its ImageNet moment.'' 

Two key missing aspects in this comparison are (i) generally accepted benchmark datasets and (ii) meaningful metrics to measure success.
Regarding (i), currently, each proposed method typically comes with its own evaluation on simulated data.
These are mostly purely synthetic and sometimes supplemented with isolated semi-synthetic case studies or evaluations on virtually the only ``gold standard'' real-world dataset the community has agreed on, a protein-signaling network published 20 years ago \citep{sachs2005causal}.
Moreover, there are growing concerns about how meaningful and generalizable reported results for causal discovery are when they are based on simulated graphs and data \citep{reisach2021beware}.
In recognition of these shortcomings, researchers have, for example, built platforms to evaluate and benchmark methods (e.g., \href{https://causeme.uv.es/}{\texttt{https://causeme.uv.es}} or \citealp{rios2025benchpress}), compiled candidate benchmark datasets based on real data \citep{chevalley2025large} or by proposing standardized procedures \citep{Geffner2022ccuite} or improved techniques \citep{herman2025unitless} for generating synthetic data, and even built physical machines with supposedly known causal structure from which measurements can be taken \citep{gamella2024causal}.

As for the evaluation metrics, one typically evaluates a distance or similarity measure on pairs of graphs, the true one and the estimated one.
Among the most intuitive and widely used metrics is the Structural Hamming Distance (SHD) \citep{Tsamardinos06SHD,perrier08SHD}, which simply counts the discrepancies between two graphs in terms of their edge set.
While there is also ambiguity in how the SHD is applied exactly, it is also unclear whether it captures ``relevant'' differences.
This motivates measuring the similarity of graphs by how much their implications differ on downstream tasks.
While extracting a causal graph can sometimes be an end goal in itself, one may then often be interested in the strength of causal connections, or the outcomes of hypothetical experiments, which can be captured by cause-effect estimation---arguably the most common and relevant downstream task of causal discovery \citep{gentzel2019case}.

The accurate estimation of causal effects from observational data is critical for reliable decision-making across various fields such as economics \citep{athey2017state, imbens2009recent}, healthcare \citep{badri2009healthcare, sanchez2022causal}, medicine \citep{castro2020causality}, and social sciences \citep{dehejia1998causal,sewell1968social}. 
Cause-effect estimation typically relies on strong assumptions about the data generating process, which in the SCM framework are naturally expressed as DAGs.
Different assumed DAGs yield different cause-effect estimands.

\citet{Peters2015SID} consequently developed the Structural Intervention Distance (SID), which essentially counts the number of intervention distributions that would be falsely inferred under a candidate graph with respect to a reference graph.
Improving upon and extending this work, \citet{henckel2024gadjid} recently proposed adjustment intervention distances that are more flexible and extend beyond DAGs also to CPDAGs.
In this work, we extend these ideas to allow for unobserved variables, leading to a graph distance for acyclic directed mixed graphs (ADMG) \citep{verma1990ADMG} making use of identification via fixing \citep{Richardson2023nested}.
We call this distance Fixing Identification Distance (FID).
Such an extension to ADMGs is relevant because incorrect or incomplete specification of latent structures can severely bias cause-effect estimation, potentially resulting in misguided interventions and policies. 

Distance measures for graphs are not only useful to evaluate causal discovery algorithms in settings where the true causal graph is known, but also to assess the sensitivity of downstream cause-effect estimation to different possible graphs more broadly.
Many algorithms for cause-effect estimation require a causal graph as input.
Since causal graphs are generally not known, the literature often alludes to either causal discovery algorithms or domain experts to provide these graphs.
However, both are often error-prone, and different algorithms/experts may give different answers.
In such settings, distance measures like SID, AID, and our FID provide insights into whether these different graphs actually make a difference for the resulting cause-effect estimates of interest.
Understanding these differences helps researchers evaluate the robustness of causal conclusions to structural assumptions, supports transparent reporting of causal inference results, and can ultimately improve decision-making.

\subsection{Related Work}
A related, but orthogonal body of work focuses on evaluating causal discovery algorithms when the ground truth graph is not or only partially known \citep{Viinikka2018intersvali,Eigenmann2020riskeval,peyrard2021laddercausaldistances,Wildberger2023KLdivcausal}.
These methods are based on empirical validation in that they rely on data to evaluate candidate graphs. 

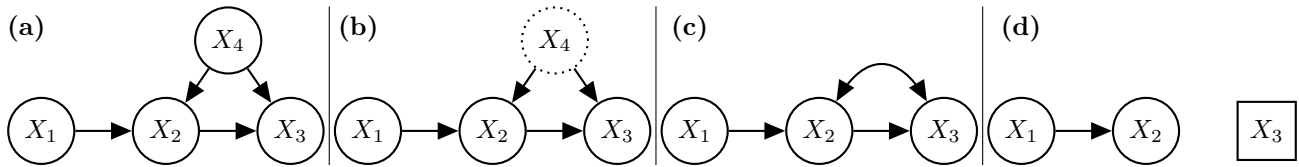
\begin{figure*}
    \textbf{(a)}\hspace{0.22\textwidth}
    \textbf{(b)}\hspace{0.22\textwidth}
    \textbf{(c)}\hspace{0.22\textwidth}
    \textbf{(d)}
    \\[-5mm]
    \begin{tikzpicture}
     \node[obs] (X1) {$X_1$};
     \node[obs, right =of X1] (X2) {$X_2$};
     \node[obs, right =of X2] (X3) {$X_3$};
     \node[obs, above = of $(X2)!0.5!(X3)$] (X4) {$X_4$};
     \edge{X1}{X2}
     \edge{X2}{X3}
     \edge{X4}{X2,X3}
  \end{tikzpicture}%
        \hfill
        \vrule
        \hfill
    \begin{tikzpicture}
     \node[obs] (X1) {$X_1$};
     \node[obs, right =of X1] (X2) {$X_2$};
     \node[obs, right =of X2] (X3) {$X_3$};
     \node[latent, above = of $(X2)!0.5!(X3)$] (X4) {$X_4$};
     \edge{X1}{X2}
     \edge{X2}{X3}
     \edge{X4}{X2,X3}
  \end{tikzpicture}%
        \hfill
        \vrule
        \hfill
   \begin{tikzpicture}
     \node[obs] (X1) {$X_1$};
     \node[obs, right =of X1] (X2) {$X_2$};
     \node[obs, right =of X2] (X3) {$X_3$};
     \edge{X1}{X2}
     \edge{X2}{X3}
     \edge[<->, bend left=60, looseness=1.6]{X2}{X3}
  \end{tikzpicture}%
        \hfill
        \vrule
        \hfill
    \begin{tikzpicture}
     \node[obs] (X1) {$X_1$};
     \node[obs, right =of X1] (X2) {$X_2$};
     \node[fixed, right =of X2] (X3) {$X_3$};
     \edge{X1}{X2}
  \end{tikzpicture}%
  \caption{Example graphs: \textbf{(a)} A DAG without latents. \textbf{(b)} A DAG with latent variable $X_4$. \textbf{(c)} An ADMG, the latent projection of (b). \textbf{(d)} The CADMG resulting from fixing node $X_3$ in (c).}\label{fig:graphs}
\end{figure*}

Instead, our graph distance is graphical, i.e., it takes as input only the graphs and no data.
Among purely graph-based distance measures, the Structural Hamming Distance (SHD) and variants thereof \citep{Tsamardinos06SHD, perrier08SHD,de2009comparison} is most common and simply counts how many pairs of nodes have different connections, i.e., types of arrows, between them in the two graphs under comparison.
The Structural Intervention Distance (SID) \citep{Peters2015SID} counts the number of differing intervention distributions implied by two different DAGs.
Most related to our work, \citet{henckel2024gadjid} propose the Adjustment Identification Distance (AID), measuring how many pairs of variables the two graphs would yield different cause-effect estimands derived from a given adjustment strategy.
Their AID applies not only to DAGs, but also to completed partially directed acyclic graphs, which represent Markov equivalence classes of DAGs.
Most recently, \citet{wahl2025separationbaseddistancemeasurescausal} proposed a separation-based distance (SD) that quantifies whether a separating set between two variables in one graph remains a separating set between the same variables in the other graph.
This separation-based measure applies specifically to DAGs and maximal ancestral graphs (MAGs) \citep{Richardson2002ancestral}, but can in principle be extended to other types of graphs with corresponding separation notions.

In this work, we stay close to the interpretation of AID by \citet{henckel2024gadjid} that directly asks: ``If I use graph $H$ to derive a consistent cause-effect estimand, will that estimand still be consistent if the true graph was $G$?''
While AID only applies in fully observed settings, we tackle the case where there may be unobserved confounders, leading to the first causal distance for ADMGs.

\subsection{Background}
In this work, we assume basic familiarity with structural causal models \citep{Pearl2000causality,peters2017elements}, including the notions of interventions, Markov equivalence classes, CPDAGs, and valid adjustment sets \citep{shpitser2012validity}.
We make use of nested Markov models, especially how causal effects in DAGs with latents can be identified via fixing within the corresponding ADMG after projecting out the latents \citep{Richardson2023nested}.
We will give a short introduction to identification via fixing, but have refer to the excellent article by \citet{Richardson2023nested} for more details due to space constraints.

\paragraph{The ``gadjid'' framework.}
\citet{henckel2024gadjid} introduced a principled framework for measuring the distance between two (CP)DAGs $G$ and $H$ based on their agreement on cause-effect estimands.
The Adjustment Identification Distance (AID) $\aid(\G, \H)$, is defined for a given cause-effect identification strategy, specifically parent adjustment, ancestor adjustment, or optimal adjustment).\footnote{Throughout this work, we denote the two graphs to be compared by $G$ and $H$, and, depending on the context, will call $G$ the ground truth/reference graph and $H$ the estimated/candidate graph.}
An \emph{identification strategy} $\cI$ is an algorithm that, given a candidate graph $\H$ and a pair of nodes $(T,Y)$ (typically interpreted as treatment and outcome), returns either a valid identifying formula for the interventional distribution $p(y\given do(t))$, or reports non-identifiability. 
A \emph{verifier} then checks if this identifying formula is correct for the reference graph $\G$. 
Formally, the identification distance is defined as
\begin{equation*}
   \aid(\G, \H,\bS) = \sum_{(T,Y)\in \bS}\mathds{1}_{\{\mathrm{incorrect}\}}(\cV(\G, \cI(\H, T, Y)))\:,
\end{equation*}
where $\bS\subseteq\{(T, Y)\in \bV\times\bV\given T\neq Y\}$ are the cause-effect pairs of interest (often all possible ones) and $\cV$ is a verifier.
Hence, $\aid$ directly counts how many cause-effect pairs would be estimated incorrectly when using the candidate graph $\H$ instead of the true graph $\G$.

\paragraph{Identification with latents via fixing.}
While there is a complete characterization of valid adjustment sets in the DAG setting \citep{shpitser2012validity} with various unique choices among them, such as parent, ancestral, and optimal adjustment sets used by \citet{henckel2024gadjid}, allowing for unobserved confounding presents a major challenge.

While \citet{Tian2002identification} established a sound and complete algorithm to identify causal effects based only on a DAG with latent variables commonly referred to as the ``ID algorithm'', it only provides one identifying formula among potentially many (when the effect is identifiable).
Hence, the ID algorithm alone does not suffice as a verifier, as it cannot check whether a given expression remains a valid estimand in a different graph with latents.
In their seminal work on nested Markov models, \citet{Richardson2023nested} introduced a powerful framework for identifying causal effects via ``fixing operations'' in the corresponding ADMGs.
This method allows for a systematic enumeration of valid expressions for a given causal effect in the presence of latent variables. 
Some terminology is required.

We denote by $\G(\bV, \bL)$ a DAG with observed nodes $\bV$ (solid circle nodes) and latent/unobserved nodes $\bL$ (dashed/dotted circle nodes).
The \emph{latent projection} of a DAG with latents $\G(\bV, \bL)$ is an acyclic directed mixed graph with nodes $\bV$ that can contain directed ($\rightarrow$) and bidirected ($\leftrightarrow$) edges---we denote it by $\G(\bV)$.
It is constructed from $\G(\bV,\bL)$ by including a directed edge $V_i\to V_j$ whenever there is a directed path from $V_i$ to $V_j$ in $\G(\bV,\bL)$ where all non-endpoints are in $\bL$; and we include a bidirected edge $V_i \leftrightarrow V_j$ whenever there is an undirected path between $V_i$ and $V_j$ in $\G(\bV,\bL)$ where all non-endpoints are non-colliders in $\bL$.
Effectively, whenever there is a latent confounding structure between $V_i$ and $V_j$, we denote this by a bidirected arrow, whereas direct influences (also mediated via latents) are still encoded via directed edges.
This gives rise to semi-Markovian models where unobserved confounding is encoded only pair-wise via bidirected edges. 
In such semi-Markovian models the observed part of the model does not satisfy the full Markov property \citep{Richardson2023nested}.
A \emph{conditional acyclic directed mixed graph} (CADMG) is an ADMG $\G(\bV,\bW)$ over two disjoint sets of nodes $\bV$, which we call \emph{random}, and $\bW$, which we call \emph{fixed} (solid square nodes) such that no node in $\bW$ has an arrowhead pointing towards it (via directed or bidirected edges).
Intuitively, we think of fixed nodes as ones that have been intervened upon with a fixed, but arbitrary value, explaining why they are freed from all influences (no arrowheads towards them).
\Cref{fig:graphs} provides examples of different graph types.

A \emph{district} in a CADMG $\G$ is a maximal bidirected-connected component and represents groups of variables affected by common latent confounding.
We write $\cD(\G)$ for the set of districts of $\G$.
We say a node $V \in \bV$ is \emph{fixable} in a CADMG $\G(\bV,\bW)$ if none of its children are in the same district as $V$.
For a fixable node $r$ in a CADMG $\G(\bV,\bW)$, we define the fixing operation 
\begin{equation*}
    \phi_r(\G(\bV,\bW)):=\G^*(\bV\backslash\{r\},\bW\cup \{r\})
\end{equation*}
where $\G^*(\bV\setminus\{r\},\bW\cup \{r\})$ is a CADMG with the subset of edges in $\G(\bV,\bW)$ that have no arrowheads at $r$.
A \emph{kernel over $\bV$ indexed by $\bW$} is a non-negative function $q_{\bV}(\cdot \given \bW): \bR^{|\bV|} \to \bR_{\ge 0}$ with $\int q_{\bV}(\x_{\bV} \given \x_{\bW})\, d\x_{\bV} = 1$ for all $\x_{\bW} \in \bR^{|\bW|}$.
If a kernel $q_{\bV}$ is Markov with respect to a CADMG $\G(\bV,\bW)$ and $r$ is a fixable node, we define \emph{the fixing operation with respect to a CADMG $\G(\bV,\bW)$} via
  \begin{equation*}
      \phi_r(q_{\bV}(\x_{\bV} \given \x_{\bW}); \G) = \frac{q_{\bV}(\x_{\bV} \given \x_{\bW})}{q_{\bV}(x_r \given \x_{\ndG{r}})}\:,
  \end{equation*}
where $\ndG{r}$ are the non-descendants of $r$ in $\G$.
If $r$ has no children in $\G$, we have
  \begin{equation*}
      \!\!\phi_r(q_{\bV}(\x_{\bV} | \x_{\bW}); \G) = \!\!\!\int\!\! q_{\bV}(\x_{\bV} | \x_{\bW}) dx_r\! =\! q_{\bV}(\x_{\bV \setminus \{r\}} | \x_{\bW})\:.
  \end{equation*}
Fixing a node without children allows us to simply marginalize it out, whereas fixing a node with descendants means diving by the kernel of that node conditioned on its non-descendants.

\citet{Richardson2023nested} show that fixing operations for different nodes commute for both graphs and kernels as long as the nodes remain fixable sequentially.
This extends to entire \emph{valid fixing sequences}, i.e., sequences of nodes $\bw = (w_1, \ldots, w_k) \in \bV$ where the nodes remain sequentially fixable in the corresponding CADMGs starting from $\G(\bV,\bW)$.
We then write $\phi_{\bw}(\G) := \phi_{w_k} \circ \cdots \circ \phi_{w_1}(\G)$ and $\phi_{\bw}(q_{\bV}; \G) := (\phi_{w_k} \circ \cdots \circ \phi_{w_1})(q_{\bV}; \G)$, where the final result is invariant to the order of the nodes in $\bw$ as long as they remain a valid fixing sequence.
This invariance is the key to obtain a whole list of valid expressions for a causal effect, as one may get different expressions for different valid fixing sequences with the ultimate guarantee of yielding the same outcome.

Next, a CADMG $\G(\bV, \bW)$ is \emph{reachable from an ADMG $\G^*(\bV \cup \bW)$} if there exists a valid fixing sequence $\bw$ for $\bW$ such that $\G = \phi_{\bw}(\G^*)$ and we then also call $\bV$ \emph{reachable in $\G^*$}.
Finally, a set of nodes $\bS \subseteq \bV$ is called \emph{intrinsic} in an ADMG $\G(\bV)$ if it is a district in any reachable graph derived from $\G(\bV)$, and we denote the set of intrinsic sets of $\G(\bV)$ by $\intr(\G(\bV))$.

This leads us to the main result of a sound and complete strategy to identify causal effects in latent variables DAGs via fixing.
\begin{theorem}[\citealp{Richardson2023nested}]\label{thm:identification}
  For an observational distribution $p(\x_{\bV})$ from a DAG model with latents $\G(\bV \cup \bL)$ and its latent projection $\G(\bV)$, $T,Y \in \bV$ different nodes, let $\Y^* = \an^{\G(\bV)_{\bV \setminus \{T\}}}_{Y} \cup \{Y\}$, where $\an^{\G(\bV)_{\bV\setminus \{T\}}}_Y$ are the ancestors of $Y$ in the graph induced by $\bV \setminus \{T\}$.
  Then if $\cD(\G(\bV)_{\Y^*}) \subset \intr(\G(\bV))$,
  \begin{equation}
      p(y \given do(t))
      = \int\!\!\! \prod_{\D \in \cD(\G(\bV)_{\Y^*})}\hspace{-5mm} \phi_{\bV \setminus \D}(p(\x_{\bV}); \G)\, d\x_{\Y^* \setminus \Y}\:.\label{eq:identification}
  \end{equation}
  If not (i.e., there exists $\D \in \cD(\G(\bV)_{\Y^*})$ that is not intrinsic in $\G(\bV)$), then $p(y \given do(t))$ is not identifiable in $\G(\bV \cup \bL)$.
\end{theorem}

\begin{figure*}
    \centering
    \includegraphics[width=\textwidth]{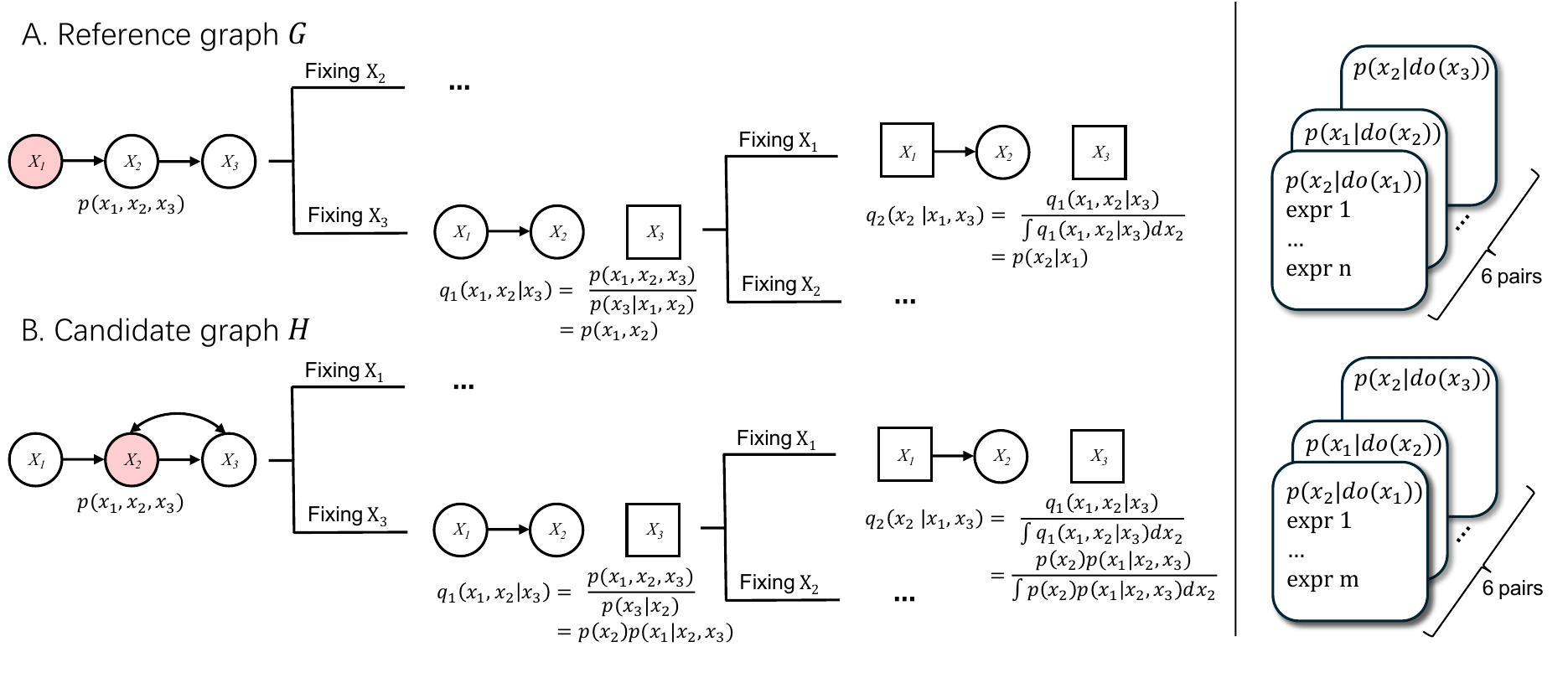}
    \caption{\textbf{Left:} Given graphs $\G$ and $\H$ and a fixed pair of nodes $(x_1, x_2)$, identification via fixing may branch according to different valid fixing sequences and lead to different symbolic expressions for $p(x_2 \given do(x_1))$ (e.g., \texttt{expr 1}, \ldots, \texttt{expr n}), where also the number of expressions can differ. The red shaded nodes indicate non-fixable variables so not all possible node sequences are valid fixing sequences. \textbf{Right:} For a set of tuples of distinct nodes, we thus obtain a list of symbolic expressions for the corresponding causal effect with respect to the two graphs. Intuitively, our FID distance then measures the aggregate/averaged overlap in the expressions for each tuple across the two graphs.}
    \label{fig:overview}
\end{figure*}

\section{Methodology}
We study the problem of measuring the discrepancy (or distance) between two causal graphs with hidden confounders represented as ADMGs $\G$ and $\H$ in a way that reflects how they (dis)agree on downstream cause-effect estimation in the sense of whether they produce equivalent formulas for $p(y \given do(t))$ for different pairs $(T,Y) \in \bV \times \bV$.
The graph $\G$ could be a known ground truth graph while $\H$ was proposed by a causal discovery algorithm, or they could both be provided by different domain experts.
We broadly follow the ``identification strategy--verifier'' framework by \citet{henckel2024gadjid} with a few important differences.

\Cref{fig:overview} provides an overview of our approach.
At the core, we exploit that in \cref{eq:identification}, for each district in the graph induced by $\Y^*$, we have a factor $\phi_{\bV \setminus \D}(\ldots)$, where we fix all nodes except for $\D$.
That means, for each factor, we may have multiple valid fixing sequences.
Hence, any combination of valid fixing sequences for each of the factors results in a valid identification formula, which are all guaranteed to be equal (under the graphical assumptions) albeit potentially being different expressions that need not be equal for any joint distribution (i.e., ones not respecting the graphical assumptions).
Our proposed identification strategy $\cI$ leverages these degrees of freedom to generate an exhaustive list of identifying formulas for all causal effects of interest.

\begin{definition}[fixing identification $\cI$]
In the setting of \cref{thm:identification}, for a pair $T,Y \in \bV$ of distinct nodes, we define $\cI(\G,T,Y)$ to yield $\emptyset$ if $p(y\given do(t))$ is not identifiable in $\G$, $p(y)$ if $Y$ is not a descendant of $T$, and the set of identifying expressions from \cref{eq:identification} for all combinations of valid fixing sequences of $\bV \setminus \D$ across the districts $\D \in \cD(\G(\bV)_{\Y^*})$.
\end{definition}
While the identification strategies of AID \citep{henckel2024gadjid} always yield a single adjustment set for each pair of nodes that can be directly compared to that of another graph, in our setting, we have variable-length lists of expressions for each pair of nodes for both graphs. 
More broadly, in the presence of latents, identification formulae cannot be broken down simply into adjustment sets, i.e., sets of nodes, but we must work directly with the identifying expressions---the estimands.
Hence, we must also adjust the notion of the verifier $\cV$ to be able to compare lists of symbolic expressions resulting from nested conditioning, building products and fractions, and marginalization of subsets of variables, see \cref{fig:overview}. 
Due to the combinatorial growth of expressions, a first step is to eliminate redundancy, i.e., expressions that are ``trivially equal'' in that they can be transformed into one another by rules of probability .
One effective approach is to reduce each expression on its own to a canonical, ``simplified'' form. 
For this, we recursively perform the following simplification steps until no pattern applies and keep a fixed order of variables for consistent results.
For this to work, we assume strictly positive densities, product spaces for all variable domains, and absolute integrability of all appearing terms. We denote this canonicalization by $\cC$.
    \begin{enumerate}[leftmargin=*,itemsep=0pt]
    \item \textbf{Conditional expansion:}
        $P(A\given B) \rightarrow \frac{P(A,B)}{P(B)}$.
    \item \textbf{Flattening and canceling fractions:}
        $\frac{\frac{A}{B}}{\frac{C}{D}}\ \mapsto\ \frac{AD}{BC}$ and
$\frac{A\cdot B}{A\cdot C}\ \mapsto\ \frac{B}{C}$ if $A\neq 0$.
    \item \textbf{Trivial marginals:}
        $\int_X  P(\ldots,X,\ldots) \to P(\ldots)$.
    \item \textbf{Canonical ordering:}
        Sums and integrals over multiple variables are sorted by the fixed pre-determined order.
        In all joint densities we also symbolically maintain the fixed variable order, e.g.,
            $P(X_2,X_1,\ldots, X_n) \to P(X_1,X_2,\ldots, X_n)$.
    \item \textbf{Consistent sorting of commutative operation:} \\
    The factors in multi-term products (divisions are considered as multiplication with the inverse) are uniquely sorted using \texttt{sympy} \citep{sympy}. 
    \end{enumerate}

This procedure turns every obtained identifying expression into a ``most simple'' form, thus de-duplicating the list of expressions per graph and node pair and allowing for fast comparison of expressions between graphs.
As acknowledged by most ``simplify'' routines in modern computer algebra systems, the uniqueness of a most canonical simple form can generally not be guaranteed for any input \citep{sympy}.
Hence, we cannot guarantee that expressions that are indeed equal for any distribution are consistently simplified to the same expression.
However, in the given setting, we still work with lists of valid expressions for comparison and have not encountered a case where symbolic comparison of equal expressions failed by the simplify-then-compare approach.

Next, we propose a symbolic verifier $\cV$ that compares canonicalized estimands across graphs. 
\begin{definition}[Symbolic Verifier $\cV$]
Given a reference graph $G$, a candidate graph $H$, a pair $(T,Y)$, and an identification strategy $\cI$, define the (multi)sets
\begin{equation*}
E_G := \mathrm{set}\big(\cC(\cI(G,T,Y))\big),\:
E_H := \mathrm{set}\big(\cC(\cI(H,T,Y))\big),
\end{equation*}
where $\emptyset$ denotes ``not identifiable.'' We write $p(y)$ for the degenerate case $Y\notin \mathrm{de}(T)$. Then the symbolic verifier can be defined as \cref{algo:verifier}.
\end{definition}
 
\begin{algorithm}
\caption{Symbolic Verifier $\cV$} \label{algo:verifier}
\begin{algorithmic}[1]
\State \textbf{Input:} reference $G$, candidate $H$, pair $(T,Y)$, identifier $\cI$, canonicalizer $\cC$
\State $E_G \gets \mathrm{set}(\cC(\cI(G,T,Y)))$
\State $E_H \gets \mathrm{set}(\cC(\cI(H,T,Y)))$
\If{$E_G=\emptyset$ and $E_H=\emptyset$} \State \Return $0$
\ElsIf{$E_G=\emptyset$ and $E_H\neq\emptyset$} \State \Return $1$
\ElsIf{$E_H=\{p(y)\}$ and $E_G\neq\{p(y)\}$} \State \Return $1$
\Else
    \State \Return $1-\frac{|E_G\cap E_H|}{|E_H|}$
\EndIf
\end{algorithmic}
\end{algorithm}

The normalization by $|E_H|$ turns the score into a \emph{proportion} of incorrectly predicted estimands. It credits partial agreement and prevents over-penalization: if only a subset of the estimands predicted by $H$ match those of $G$, the value decreases linearly with the matched fraction. The score lies in $[0,1]$ and reaches $1$ in the worst case, e.g., when $H$ asserts identifiability but none of its estimands agree with $G$.

Accordingly, $\cV$ can be read as the \emph{false-positive rate} of the candidate $H$ relative to the reference $G$---the fraction of predicted estimands unsupported by $G$. Swapping the roles of $G$ and $H$ yields the complementary \emph{false-negative rate}, i.e., the fraction of reference estimands not recovered by $H$. We now aggregate one or both of these per-pair scores over a set $\bS$ of treatment-outcome pairs to define a graph-level distance between ADMGs.

\begin{definition}[Fixing Identification Distance (FID)]
\label{def:FID}
We define our fixing identification distance via
\begin{equation}
\label{eq:distance}
    \aid(G,H;\bS) \;=\; \sum_{(T,Y)\in \bS} \cV(G,H;T,Y)\:,
\end{equation}
and its normalized version as
\begin{equation}
\label{eq:distance-normalized}
\aidbar(G,H;\bS)
:= \frac{1}{|\bS|}\, \aid(G,H;\bS)\:.
\end{equation}
Since (normalized) FID are non-symmetric, we also introduce a symmetrized version
\begin{equation}
    \aidsym(G_1,G_2;\bS)
    \;=\; \frac{1}{2}\Big(\aid(G_1,G_2;\bS) + \aid(G_2,G_1;\bS)\Big).
\end{equation}
and denote its normalized version by $\aidsymbar(G_1,G_2;\bS)$.
\end{definition}
As aggregated proportions, unlike edit-based graph scores, FID is generally continuous.

FID is also a \emph{pre-metric}: it is nonnegative and vanishes on the diagonal, but it need not be symmetric (except for the symmetrized version) nor satisfy the triangle inequality.

\section{Empirical Results}
We first illustrate the behavior of FID qualitatively and quantitatively on simulated data and then compare it with existing metrics. Since unlike SID and AID, FID operates on ADMGs, we can only meaningfully compare to a separation-based distance computed on MAGs. Unless noted otherwise, we report the normalized score $\aidsymbar$ in \cref{def:FID}.

\subsection{Graph Generation}
We sample graphs using an Erd\"{o}s--R\'enyi (ER) design. Given $n_{\text{vars}}$ observed variables:
\begin{enumerate}[leftmargin=*, itemsep=0pt]
\item \textbf{DAG backbone.} Draw a random topological order $\pi$ of the nodes. For each ordered pair $(i,j)$ with $\pi(i)<\pi(j)$, include a directed edge $i\to j$ independently with probability $p_{\text{dir}}$.
\item \textbf{Overlay latent confounding.} Add exactly $n_{\mathrm{bi}}$ bidirected edges by sampling that many unordered node pairs without replacement.
\end{enumerate}
We use a grid over $(p_{\text{dir}}, n_{\mathrm{bi}})$ and draw the same number of graphs per setting. In our configuration we set $n_{\text{vars}}=5$, sweep $p_{\text{dir}}\in\{0.2,0.3,0.4,0.5,0.6,0.7,0.8\}$ and $n_{\mathrm{bi}}\in\{1,2,3\}$, and generate $20$ graphs per combination, for a total of $7\times 3\times 20=420$ ADMGs.
For $\bS$ we use all possible tuples of distinct nodes and aggregate results over graph pairs by averaging.

\begin{figure*}
  \centering
    \includegraphics[width=0.32\textwidth]{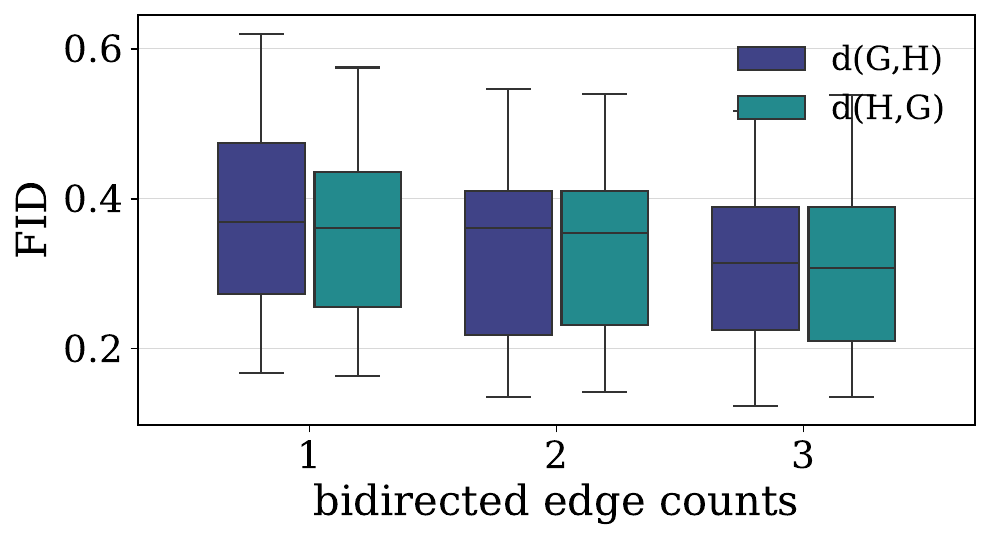}
  \hfill
    \includegraphics[width=0.32\textwidth]{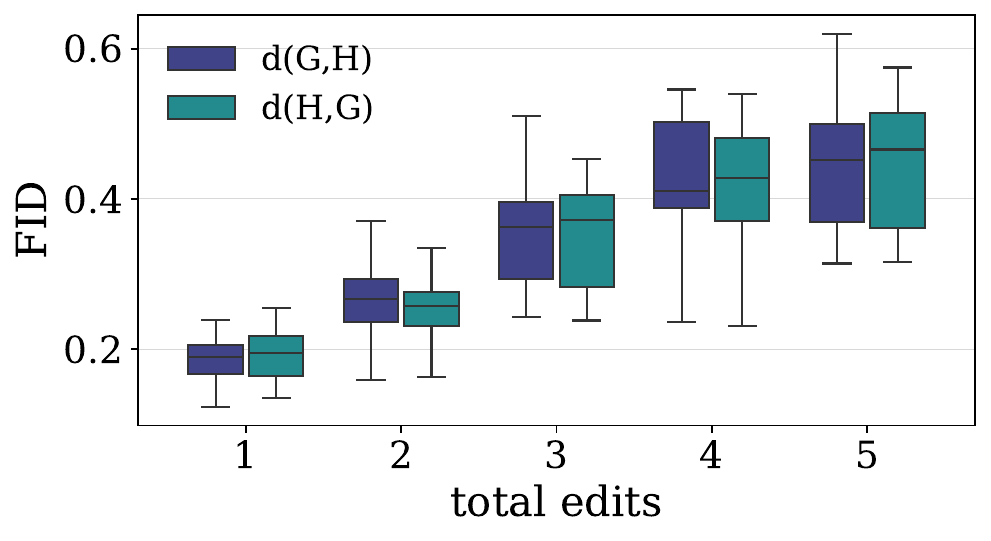}
  \hfill
    \includegraphics[width=0.32\textwidth]{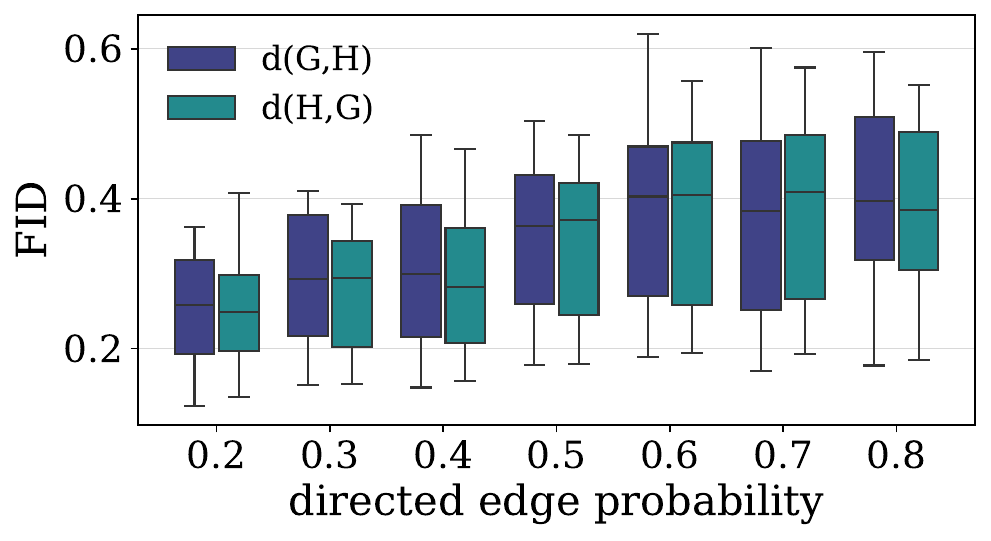}
  \caption{Boxplots show $\aidbar(G,H;\bS)$ (blue) and $\aidbar(H,G;\bS)$ (teal) under different aggregations as we vary the bidirected edge counts (left), total number of edits (middle), and directed edge probability (right).}
  \label{fig:metrics}
\end{figure*}

\subsection{FID Results}
\label{sec:empirical-metrics}

To obtain a candidate ADMG $H$, we apply $k \in \{1,2,3,4,5\}$ edits to a given $G$ chosen from the following types: reverse a directed edge (\emph{reverse dir}), turn a directed edge into a bidirected edge (\emph{di to bi}), a bidirected one into a directed one (\emph{bi to di}), add a directed edge (\emph{add dir}), delete directed edge (\emph{del dir}), add bidirected edge (\emph{add bi}), delete bidirected edge (\emph{del bi}).
Edits are applied sequentially (skipping ones that would create cycles).
The three panels in \cref{fig:metrics} show different aggregations of the corresponding distances: (i) grouped by the number of bidirected edges in $G$ and pool results over all $k$ and $p_{dir}$; (ii) pool all $G$ and group by total edits $k$; (iii) group by the $p_{dir}$ used for $G$ and pool all $k$ and $n_{bi}$. 
As expected, FID clearly increases with the number of total edits.
The original density of directed edges also tends to increase FID, which may be explained by an edit in a dense graph having more wide ranging consequences for resulting cause-effect estimands than in sparse graphs.

\begin{figure}
    \centering
    \includegraphics[width=0.8\linewidth]{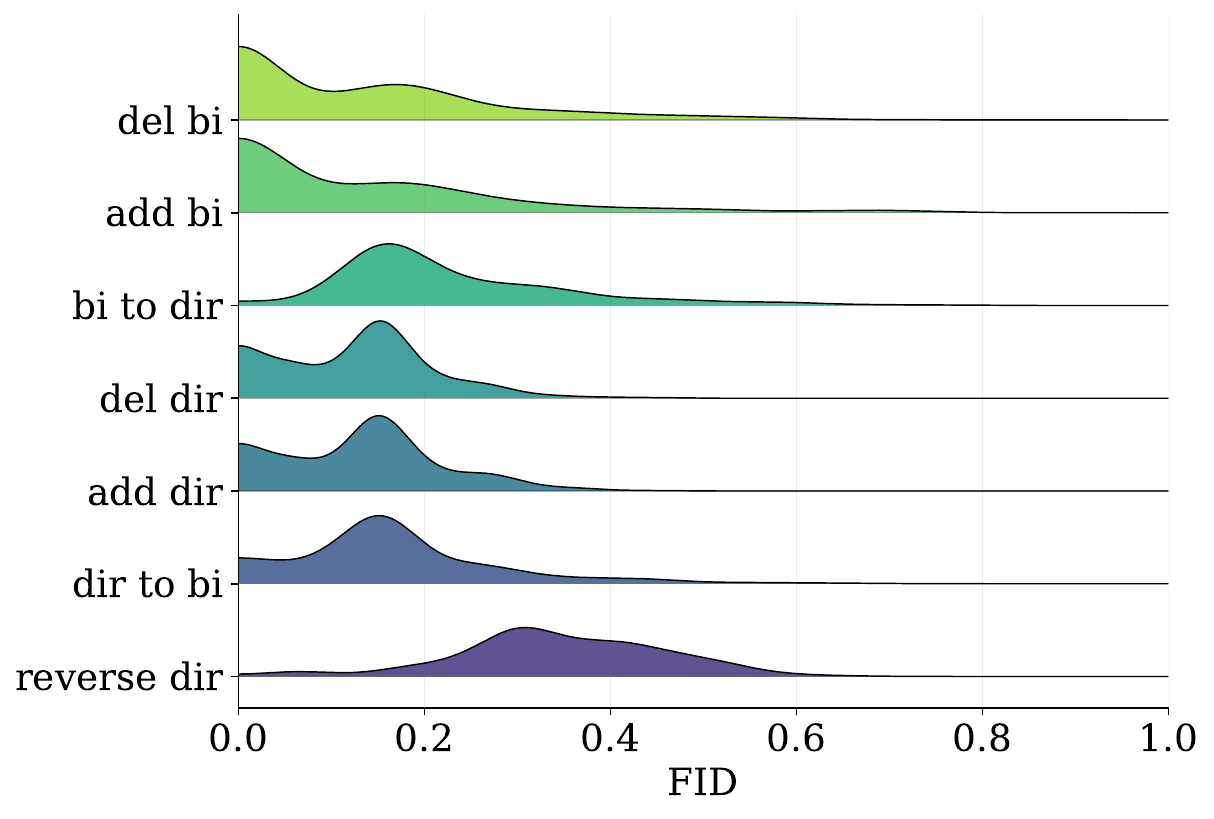}
    \caption{FID ($\aidsymbar$) distributions for the different edit types.}
    \label{fig:histogram}
\end{figure}

We next isolate single-edit effects according to the different types of edits.
Starting from $G$, we apply exactly one edit to obtain $H$ and evaluate $\aidsymbar(G,H;\bS)$. \Cref{fig:histogram} shows the distribution of distances for different edit types.
While all of these edits would count equally in a simple SHD metric, there are substantial differences in FID.
Most notably, reversing an edge simultaneously changes ancestor relations and can create/remove v-structures or open/close back-door paths.
Even one such edit can thus lead to a large FID.
We can similarly hypothesize that changing types of edges (\emph{dir to bi}, \emph{bi to dir}) or removing/adding directed edges lead to substantial changes in causal ordering and thus have a sizable impact on cause-effect estimands.
The addition or deletion of bidirected edges (introducing/removing latent confounding) on average leads to the lowest FID; however, the distribution has a heavy tail, i.e., also these edits can lead to substantial differences in downstream cause-effect estimation captured by FID.

\begin{figure}
    \includegraphics[width=\linewidth]{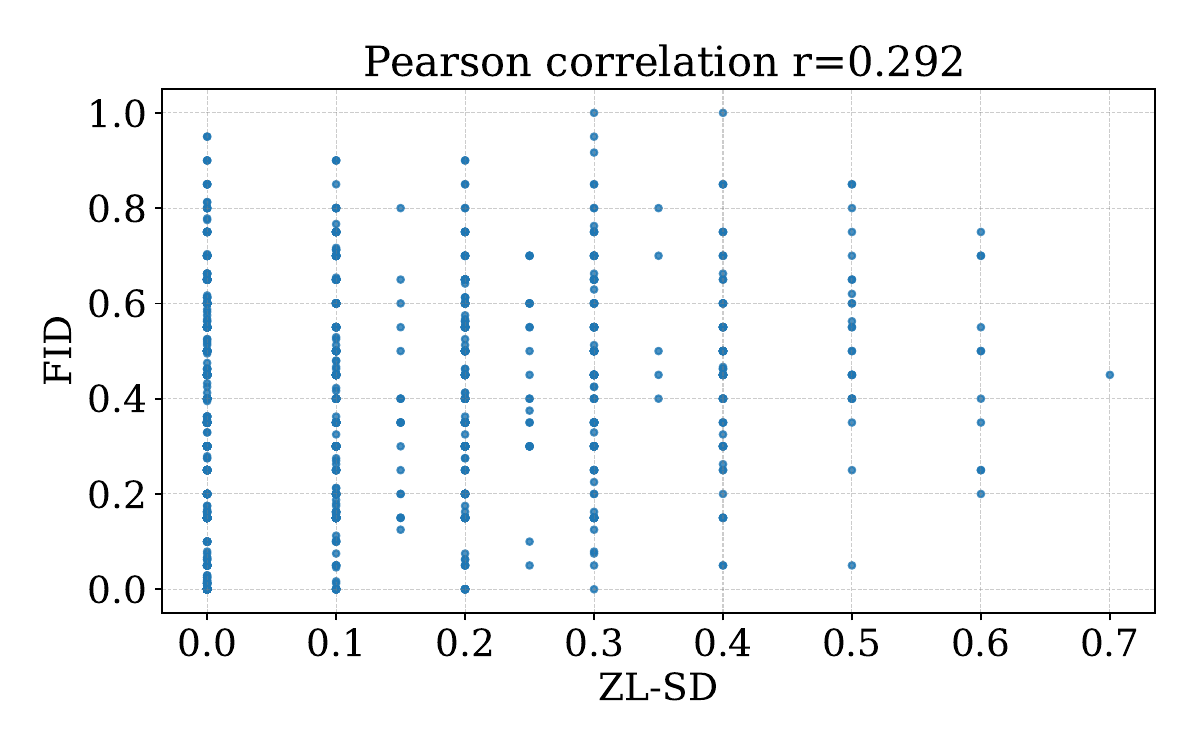}\\
    \includegraphics[width=\linewidth]{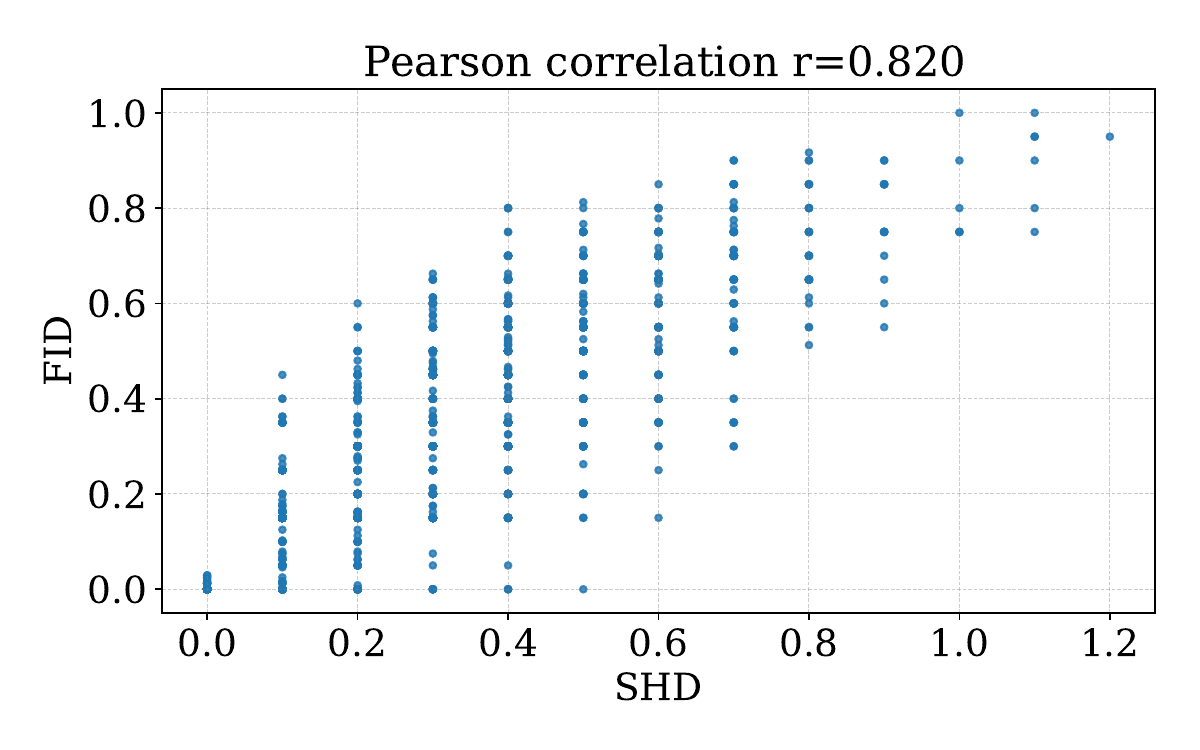}
  \caption{Comparison of our FID ($\aidsymbar$) with SHD and ZL-SD on MAGs. FID shows positive correlation with both, but still adds new information beyond existing metrics.}
  \label{fig:comparison}
\end{figure}

\subsection{Comparison with other graph distances}
\label{sec:comparison}
We compare our estimand-based distance to two structural baselines that can be applied in the presence of latent confounding: the separation-based distance (we use the normalized ZL variant) and Structural Hamming Distance.
Because SD is defined on MAGs, we convert our ADMGs to MAGs as follows. 
Each reference ADMG $G$ is mapped to a Markov-equivalent MAG $\tilde G$ using the procedure of \citet{hu2020mag2admg}; if a Markov-equivalent MAG does not exist, we discard the instance. 
We then generate perturbed ADMGs by applying $k\in\{1,\dots,7\}$ random legal edits (as in \cref{sec:empirical-metrics}), convert the resulting graphs to MAGs, and obtain $(\tilde G,\tilde H)$. The conversion can collapse distinct ADMG perturbations to the same MAG, so the number of MAG discrepancies may differ from the number of ADMG edits.

For each pair $(\tilde G,\tilde H)$ we compute SD$(\tilde G,\tilde H)$, SHD$(\tilde G,\tilde H)$, and our symmetric normalized distance $\aidsymbar(\tilde G, \tilde H ;\bS)$. \Cref{fig:comparison} shows scatter plots across all graphs and edit levels.

The comparison with ZL-SD shows that FID and ZL-SD indeed capture different aspects of graph differences albeit being weakly correlated (Pearson $r=0.292$).
SD aggregates, which (m-)separations change; many distinct edit patterns collapse to the same SD value, producing vertical stripes.
Small SD moves that open/close a confounded path can cause large jumps in FID, while large SD moves can be redundant for the fixing formulas and leave FID small.
Hence, FID and ZL-SD are complementary causal graph distances.

The comparison with SHD shows a stronger correlation (Pearson $r=0.820$), but the scatter plot still shows substantial variation within given SHD bins, highlighting that FID is not equivalent to or perfectly predicted by SHD.

\section{Conclusion}

\paragraph{Summary.}
We introduced the \emph{Fixing Identification Distance} (FID), an estimand-level distance measure for ADMGs that compares graphs by the causal answers they entail to cause-effect estimation queries under latent confounding.
FID combines fixing-based identification with symbolic canonicalization and a symbolic verifier to (i) determine identifiability and (ii) compare the resulting functionals across treatment-outcome pairs, yielding a fractional, task-aligned pre-metric. A symmetric variant supports settings without a designated ground truth. This reframes graph evaluation around the downstream causal queries that many researchers ultimately care about.

Empirically, FID behaves coherently under controlled perturbations: it increases roughly monotonically with the number of edits and with directed-edge density, and its sensitivity varies systematically by edit type. Compared to structural baselines, FID aligns strongly with SHD (edit cardinality) but only weakly with separation-based distances on MAGs, consistent with the non-linear link between (m-)separation changes and identifiability. Because DAGs are ADMGs without bidirected edges, FID also subsumes fully observed settings.
Overall, FID provides a principled causal distance measure on ADMGs that can be used to evaluate causal discovery algorithms as well as assess the sensitivity of cause-effect estimation to differences in assumed/proposed graphs.

\paragraph{Limitation and future work.}
Unlike identification on fully observed graphs, fixing operation based on ID algorithm \citep{Shpitser2006ID} for identification has an exponential complexity $O(\text{poly(n)}\times4^b)$ in graph binary width $b$ \citep{Shpitser2012complexity}. However, running fixing for all possible fixing sequences could be as expensive as $O((n/e)^n\sqrt{n})$. 
Combining both, this eventually results in the complexity of the proposed graph distance growing super-exponentially with the graph size $n$. 
The current direct implementation depends on a computer algebra system for symbolic simplification and comparison, which impedes the use of modern hardware accelerators and is a limiting factor for computational efficiency. This computational obstacle currently limits the scalability to very large graphs. 
In practice, after canonicalization only few unique identifying expressions remain. Therefore, we suspect that early pruning of expressions by clever application of graphical criteria along different fixing sequences together with parallelization could substantially speed up the current method and scale it to larger graphs.
Finally, our canonicalization is sound but not complete---algebraically distinct expressions may still be equivalent under nested-Markov constraints---so FID may technically under-merge some estimands.
Extending equivalence checking (e.g., richer rewrite systems or SMT-aided verification), covering multi-treatment and path-specific effects, and benchmarking on semi-synthetic/real tasks are natural extensions for future work.

\subsection*{Acknowledgments}

This work was funded by Helmholtz Association’s Initiative and Networking Fund through Helmholtz AI.
This work has been supported by the German Federal Ministry of Education and Research (Grant: 01IS24082).

\bibliographystyle{plainnat}
\bibliography{bib.bib}
\clearpage

\appendix
\thispagestyle{empty}

\onecolumn
\aistatstitle{Appendix}

\section{Additional Background}

\paragraph{Basic graph notation.}
For a graph $G$ over nodes $\bV$ and $X\in\bV$, $\pa_G(X)$, $\ch_G(X)$, $\de_G(X)$, $\an_G(X)$, and $\nd_G(X)$ denote the parents, children, descendants, ancestors, and non-descendants of $X$ in $G$, respectively. 
\begin{align*}
\pa_G(X) &:= \{\,Y\in\bV\setminus\{X\} : Y \to X \text{ is an edge in } G\,\},\\
\ch_G(X) &:= \{\,Y\in\bV\setminus\{X\} : X \to Y \text{ is an edge in } G\,\},\\
\an_G(X) &:= \{\,Y\in\bV : \text{there exists a directed path } Y \rightsquigarrow X\,\},\\
\de_G(X) &:= \{\,Y\in\bV : \text{there exists a directed path } X \rightsquigarrow Y\,\},\\
\nd_G(X) &:= \bV\setminus \{\de_G(X) \cup X\}.
\end{align*}
A (directed) \emph{path} in $G$ is a sequence of distinct nodes $v_0,\dots,v_k$ such that
$v_i \to v_{i+1}$ is a directed edge of $G$ for all $i$. By convention, a node $v_i$ is neither an ancestor, nor a descendant,
nor a non-descendant of itself. If $G$ contains bidirected edges ($\leftrightarrow$), the \emph{siblings} of $X$ are its bidirected neighbors, and the \emph{district} $\dis_G(X)$ is the bidirected-connected component containing $X$. In DAGs, the (probabilistic) \emph{Markov blanket} of $X$ is $\mb_G(X)=\pa_G(X)\cup\ch_G(X)\cup\big(\pa_G(\ch_G(X))\setminus\{X\}\big)$. In CADMGs used by nested Markov models, the \emph{Markov blanket} that appears in the nested factorization is
$\mb_G(X)=\big(\dis_G(X)\setminus\{X\}\big)\cup\pa_G\big(\dis_G(X)\big)$.

\paragraph{Induced subgraph.}
Let $G$ be a (mixed) graph on vertex set $\bV$ with edge set $E$ (edges may be directed $\to$ or bidirected $\leftrightarrow$). For $A\subseteq\bV$, the \emph{induced subgraph} of $G$ on $A$, denoted $G_A$, is the graph whose
\begin{equation*}
\text{vertex set is } A,\qquad
\text{edge set is } E_A \;=\; \{\, e\in E : \text{both endpoints of } e \text{ lie in } A \,\},
\end{equation*}
with each edge in $E_A$ retaining the same type and orientation it has in $G$.  
Equivalently, $G_A$ is obtained by deleting all vertices in $\bV\setminus A$ and then removing any edges incident to the deleted vertices.

\paragraph{Graph classes.}
A \emph{DAG} has only directed edges ($\rightarrow$) and no directed cycles. An \emph{ADMG} (acyclic directed mixed graph) has directed and bidirected edges but no directed cycles; bows ($X\rightarrow Y$ together with $X\leftrightarrow Y$) are allowed. A \emph{CADMG} $G(\bV,\bW)$ partitions nodes into random $\bV$ and fixed $\bW$ and contains no arrowheads into $\bW$. A \emph{CPDAG} (completed partially directed acyclic graph) compactly represents a Markov equivalence class of DAGs \citep{Andersson1997CPDAG}: directed edges are oriented the same way in every DAG in the class, while undirected edges indicate reversible orientations. Two DAGs are \emph{Markov equivalent} if and only if they share the same skeleton and v-structures. An \emph{ancestral graph} is a mixed graph with (i) no directed cycles; (ii) no arrowhead into an ancestor (e.g., if $X\leftrightarrow Y$ then neither $X\to Y$ nor $Y\to X$ is present); and (iii) if $X-Y$ is undirected, then $X$ and $Y$ have no incident arrowheads. A \emph{MAG} (maximal ancestral graph) \cite{Richardson2002ancestral} is an ancestral graph which is maximum if and only if for every pair of nonadjacent vertices $X$, $Y$, there exists some set $\bZ \subseteq \bV\setminus \{X, Y\}$ such that $X$ and $Y$ are m-separated given $\bZ$. Intuitively, in a MAG every missing edge corresponds to at least one conditional independence; adding any edge would change the independence model. 

\paragraph{Separation.}
In DAGs, $d$-separation characterizes conditional independencies.
For mixed graphs (ancestral graphs, ADMGs), the corresponding notion is \emph{m-separation}:
A path is m-connecting given $\bZ$ if every non-collider on the path is \emph{not} in $\bZ$ and every collider on the path is an ancestor of some node in $\bZ$; otherwise, the endpoints are m-separated given $\bZ$.

\paragraph{Latent projection and districts.}
Given a DAG $G(\bV\cup\bL)$ with latents $\bL$, its \emph{latent projection} onto $\bV$ is the ADMG obtained by (i) adding $X\to Y$ if there is a directed path $X\rightsquigarrow Y$ whose non-endpoints are latent, and (ii) adding $X\leftrightarrow Y$ if there is a path between $X$ and $Y$ whose non-endpoints are non-colliders and latent. Bidirected-connected components in an (C)ADMG are called \emph{districts}.

\paragraph{Markov equivalence of ADMG and MAG.}
Two (mixed) graphs on the same vertices are \emph{Markov equivalent} if they imply exactly the same set of conditional independencies by m-separation. Every ADMG has a Markov-equivalent MAG obtained by the projection in \cref{alg:admg-to-mag} \citep{hu2020mag2admg}. 

\paragraph{Tails.}
For $\mathbf{A}\subseteq \bV$, write $G_\mathbf{A}$ for the induced subgraph on $\mathbf{A}$. For a set $S$, let $\an_G(S)$ be all ancestors of nodes in $S$ (including $S$). The \emph{district} of $v$ in $G_\mathbf{A}$, denoted $\dis_{G_\mathbf{A}}(v)$, is the bidirected-connected component containing $v$ in $G_\mathbf{A}$. Define the \emph{tail} \citep{richardson2014factorizationcriterionacyclicdirected} of a single vertex $v$ by
\begin{equation*}
\mathrm{tail}_G(v)\ :=\ \big(\dis_{G_{\{v\}\cup \an_G(\pa_G(v))}}(v)\setminus\{v\}\big)\ \cup\ \pa_G\!\big(\dis_{G_{\{v\}\cup \an_G(\pa_G(v))}}(v)\big).
\end{equation*}
Intuitively, $\mathrm{tail}_G(v)$ collects variables that, in the ancestor subgraph around $v$’s parents, share $v$’s bidirected component or parent of that component. In the projected MAG these become the parents of $v$.

\paragraph{Heads.}
A set $\mathbf{H}\subseteq \bV$ is a \emph{head} \citep{richardson2014factorizationcriterionacyclicdirected} if it is barren (no member has a descendant within $\mathbf{H}$ other than itself) and $\mathbf{H}$ lies in a single district of the ancestor subgraph $G_{\an_G(\mathbf{H})}$. When $|\mathbf{H}|=2$, say $\mathbf{H}=\{v,w\}$, this exactly means $v$ and $w$ lie in the same district of $G_{\an_G(\{v,w\})}$. In the projection, such pairs correspond to bidirected edges $v\leftrightarrow w$ in the MAG.

\section{FID on DAGs and CPDAGs}

A DAG is an ADMG without bidirected edges, so FID applies directly. 
In a DAG, all effects are identifiable; this results also apply to the fixing operation. 
In a fully observed DAG, every interventional distribution $p(y\mid do(t))$ is identifiable. Moreover, every node forms a singleton district; hence, all nodes are fixable and reachable.
While both AID and FID are identification-based, AID depends on a chosen adjustment strategy (e.g., parent, ancestor, or optimal adjustment), whereas FID derives the fixing-based estimand implied by the graph. Consequently, they may disagree for a given pair $(T,Y)$.

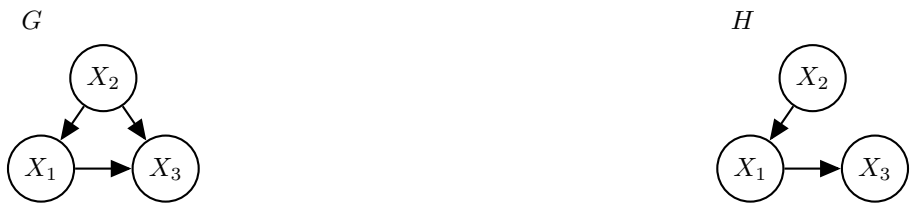
\begin{figure}
\centering
\subcaptionbox{}[0.45\linewidth]{
\begin{tikzpicture}[baseline=(current bounding box.north)]
  \node[obs] (X1) {$X_1$};
  \node[obs, right=of X1] (X3) {$X_3$};
  \node[obs, above=of $(X1)!0.5!(X3)$] (X2) {$X_2$};
  \node[anchor=south west, xshift=2pt, yshift=2pt]
  at (current bounding box.north west) {\textbf{$G$}};
  \edge{X1}{X3}
  \edge{X2}{X1}
  \edge{X2}{X3}
\end{tikzpicture}%
}
\hfill
\subcaptionbox{}[0.45\linewidth]{
\begin{tikzpicture}[baseline=(current bounding box.north)]
  \node[obs] (X1) {$X_1$};
  \node[obs, right=of X1] (X3) {$X_3$};
  \node[obs, above=of $(X1)!0.5!(X3)$] (X2) {$X_2$};
  \node[anchor=south west, xshift=2pt, yshift=2pt]
  at (current bounding box.north west) {\textbf{$H$}};
  \edge{X1}{X3}
  \edge{X2}{X1}
\end{tikzpicture}%
}
\caption{Example illustrating the discrepancy between AID and FID.}
\label{fig:AID<FID}
\end{figure}

\paragraph{Example.}
In \cref{fig:AID<FID}, let $T=X_1$ and $Y=X_3$, the identifier (parent and ancestor adjustment) of the AID provides the same expression 
\begin{equation*}
\cI(G;x_1,x_3)=\int p(x_3\given x_1,x_2)p(x_2)dx_2=\cI(H;x_1,x_3)\:.
\end{equation*}
Therefore, the verifier of AID would give a zero distance. 
By contrast, fixing algorithms disagree on these two graphs. On graph $G$, $Y^*=\{X_2,X_3\}$ and the identifier of FID gives the following functional (before converting to symbolic form)
\begin{equation*}
    p(x_3\given do(x_1))=\int \phi_{x_1,x_3}(p(\mathbf{x_V};G))\phi_{x_1,x_2}(p(\mathbf{x_V};G))dx_2.
\end{equation*}
While on graph $H$, $Y^*=\{X_3\}$ and the functional providing by identifier of FID is
\begin{equation*}
p(x_3\given do(x_1))=\phi_{x_1,x_2}(p(\mathbf{x_V};H)).
\end{equation*}

These two estimands are generally \emph{not} algebraically equal when evaluated under distributions Markov to $G$ (in $G$, $X_2 \not\!\perp\!\!\!\perp X_1$ and $X_2 \not\!\perp\!\!\!\perp X_3 \given X_1$), so FID records a disagreement for $(T,Y)$.

By contrast, depending on the adjustment strategy, AID may output the \emph{same} adjustment expression for both graphs (e.g., a strategy that always adjusts for $X_2$), leading to zero AID on this pair. FID is therefore more sensitive to the minimal fixing-based estimand implied by each DAG: it detects that in $G$ an adjustment is required, whereas in $H$ it is not.

\captionsetup[subfigure]{%
  labelformat=empty,
  position=top,
  justification=raggedright,
  singlelinecheck=false,
  font=bf,
  skip=0pt
}

The DAG case considered above extends naturally to CPDAGs, which encode Markov equivalence classes of DAGs. Since different consistent orientations within a CPDAG may induce different identifying functionals for a given $(T,Y)$, a single scalar FID is no longer well-defined. Following the SID treatment \citep{Peters2015SID}, we therefore evaluate FID \emph{over all DAG extensions} of each CPDAG and summarize the result as a range. Concretely, for CPDAGs $C_1,C_2$, we report the minimum and maximum normalized directional FID attained by any pair of DAGs $G\in[C_1]$, $H\in[C_2]$. The interval collapses to a point when both inputs are DAGs. In practice, exact enumeration is feasible for small graphs; for larger CPDAGs, one can sample consistent orientations to approximate the bounds.

Let $C$ be a CPDAG and $[C]$ the set of DAGs that are Markov equivalent to $C$ (its DAG extensions). For a set of ordered treatment-outcome pairs $\bS\subseteq\{(T,Y)\in\bV\times\bV:T\neq Y\}$, define the \emph{directional FID range} between two CPDAGs $C_1,C_2$ as
\begin{equation*}
\mathrm{FID}^{\mathrm{range}}_{\cI}(C_1,C_2;\bS)
\;:=\;
\big[\;\underline d,\;\overline d\;\big]
\quad\text{with}\quad
\begin{aligned}
\underline d &:= \min_{G\in[C_1]}\;\min_{H\in[C_2]}\; \aidbar(G,H;\bS),\\
\overline d  &:= \max_{G\in[C_1]}\;\max_{H\in[C_2]}\; \aidbar(G,H;\bS),
\end{aligned}
\end{equation*}
where $\aidbar$ is the normalized directional FID (\cref{eq:distance-normalized}). The interval contracts to a point if both graphs are DAGs. A symmetric variant follows by replacing $\aidbar$ with its symmetrization $ \aidsymbar$ (\cref{def:FID}). A detailed algorithm can be found in \cref{alg:cpdag-fid}. 

\begin{algorithm}
\caption{CPDAG-FID range (exact or sampled)}
\label{alg:cpdag-fid}
\begin{algorithmic}[1]
\State \textbf{Input:} CPDAGs $\cC_1,\cC_2$, pair set $\bS$, identifier $\cI$, canonicalizer $\cC$, mode $\in\{\textsc{Exact},\textsc{Sample}\}$, budget $B$
\If{\textsc{Exact}}
  \State $\mathcal{G}\gets[\cC_1]$ 
  \State $\mathcal{H}\gets[\cC_2]$
\Else
  \State $\mathcal{G}\gets$ $B$ DAGs sampled uniformly from $[\cC_1]$
  \State $\mathcal{H}\gets$ $B$ DAGs sampled uniformly from $[\cC_2]$
\EndIf
\State $\underline d \gets +\infty$, \ $\overline d \gets 0$
\For{$G\in\mathcal{G}$}
  \For{$H\in\mathcal{H}$}
    \State $s \gets \aidbar(G,H;\bS)$ 
    \State $\underline d \gets \min(\underline d, s)$; \ $\overline d \gets \max(\overline d, s)$
  \EndFor
\EndFor
\State \textbf{return} $[\underline d,\overline d]$
\end{algorithmic}
\end{algorithm}

\section{Experiment Setup}

\subsection{Hardware}

Our experiments were conducted on a scheduled cluster with heterogeneous CPU nodes, and the most common types are Intel Xeon 6248r and Intel Xeon Gold 6142. For a single job, we utilized 64 CPU cores and 64GB of memory. 

\subsection{ADMG\texorpdfstring{$\;\to\;$}{->}MAG Projection}
\label{sec:admg-to-mag}

In \cref{sec:comparison}, we convert an ADMG $G$ into a Markov equivalent MAG $G_m$ by reconstructing exactly those
directed and bidirected edges that are determined by the \emph{tails} of single vertices and by \emph{heads} of size two. In particular (Lemma~3.7 in \citealp{hu2020mag2admg}), for any vertices $v,w$: (i) $w\to v$ is present in $G_m$ if and only if $w\in \mathrm{tail}_G(v)$; (ii) $v\leftrightarrow w$ is present in $G_m$ if and only if $\{v,w\}$ is a head in $G$ (equivalently, $v$ and $w$ are in the same district inside the ancestor subgraph of $\{v,w\}$). The procedure below computes these objects in polynomial time and yields a MAG that is Markov equivalent to the input ADMG (Table 5 in \citealp{hu2020mag2admg}).

\begin{algorithm}[H]
\caption{ADMG$\;\to\;$MAG projection (Algorithm~2 in \citealp{hu2020mag2admg})}
\label{alg:admg-to-mag}
\begin{algorithmic}[1]
\State \textbf{Input:} ADMG $G=(V,E)$ \hfill (access to $\pa_G(\cdot)$, $\an_G(\cdot)$, districts)
\State \textbf{Output:} MAG $G_m=(V,E_m)$ Markov equivalent to $G$
\State Initialize $G_m$ with vertex set $V$ and no edges: $E_m \gets \emptyset$
\For{each $v\in V$} \label{line:loop-v}
  \State $A_v \gets \{v\}\cup \an_G(\pa_G(v))$ 
  \State $D_v \gets \dis_{G_{A_v}}(v)$ 
  \State $\mathrm{tail}_G(v) \gets \big(D_v \setminus \{v\}\big)\ \cup\ \pa_G(D_v)$
  \For{each $w\in \mathrm{tail}_G(v)$}
    \State add edge $w\to v$ to $E_m$
  \EndFor
\EndFor
\For{each unordered pair $\{v,w\}\subseteq V$ \textbf{with} $v\notin \an_G(w)$ \textbf{and} $w\notin \an_G(v)$ \textbf{and} $v,w$ in the same district of $G$} \label{line:loop-pairs}
  \State $A_{vw} \gets \{v,w\}\cup \an_G(\{v,w\})$
  \State $D_{vw} \gets \dis_{G_{A_{vw}}}(v)$
  \If{$w \in D_{vw}$}
    \State add edge $v\leftrightarrow w$ to $E_m$
  \EndIf
\EndFor
\State \Return $G_m$
\end{algorithmic}
\end{algorithm}

\section{Additional Experiment Results}

\paragraph{Tabulated summaries.}
\Cref{tab:1,tab:2,tab:3,tab:4} report the \emph{directional}, normalized FID $\aidbar(\cdot,\cdot;\bS)\in[0,1]$ as precise counterparts to the box/violin plots in \cref{sec:empirical-metrics}. Higher values mean a larger fraction of treatment-outcome estimands disagree after canonicalization.

\emph{Single-edit breakdown (\cref{tab:1,tab:2}).} For each $(n_{\mathrm{bi}},p_{\mathrm{dir}})$ setting, we apply exactly one legal edit to a reference ADMG $G$ to obtain $H$ and average $\aidbar$ over graphs and edit locations. \Cref{tab:1} lists $\aidbar(G,H;\bS)$ and \cref{tab:2} lists the reverse direction $\aidbar(H,G;\bS)$. The per-edit-type ordering matches \cref{fig:histogram}: \textit{reverse dir} and \textit{dir to bi} are largest on average, while \textit{del bi} and \textit{add bi} are smallest, with slight directional asymmetry between \cref{tab:1} and \cref{tab:2}.

\emph{Edit-count trends (\cref{tab:3,tab:4}).} These tables aggregate $\aidbar$ by the number of successful edits $k\in\{1,\dots,5\}$ (columns), again for both directions. They complement the middle panel of \cref{fig:metrics}: values reflect averages over random edit types at each $k$, so exact numbers need not be strictly monotone across $k$ within every $(n_{\mathrm{bi}},p_{\mathrm{dir}})$ cell. Differences between \cref{tab:3} and \cref{tab:4} quantify the inherent directionality of FID.

\emph{Pooled edit-type means (\cref{tab:breakdown}).} Finally, we pool across all $(n_{\mathrm{bi}},p_{\mathrm{dir}},k)$ and report the mean and standard error of $\aidbar$ per edit type. The ranking mirrors the ridgeline plot in \cref{fig:histogram}: edits that reverse a direction or introduce confounding (\textit{reverse dir}, \textit{dir to bi}) have the largest average impact on estimands; removing bidirected edges (\textit{del bi}) has the smallest.

\begin{table}
\centering
\rowcolors{2}{gray!10}{white}
\caption{Single-edit FID, $\G\to \H$ --- mean $\aidbar$ by $(n_{\mathrm{bi}},p_{\mathrm{dir}})$ and edit type.} \label{tab:1}
\begin{tabular}{cc|ccccccc}
\toprule
\textbf{bi} & \textbf{dir} & \textbf{reverse} & \textbf{di to bi} & \textbf{add dir} & \textbf{del dir} & \textbf{bi to di} & \textbf{add bi} & \textbf{del bi} \\ \midrule
1 & 0.2 & 0.3278 & 0.1556 & 0.1675 & 0.1722 & 0.1817 & 0.0875 & 0.0350 \\
1 & 0.3 & 0.3368 & 0.1895 & 0.1725 & 0.1855 & 0.2150 & 0.1475 & 0.0863 \\
1 & 0.4 & 0.3550 & 0.1975 & 0.1625 & 0.1425 & 0.2500 & 0.1050 & 0.1200 \\
1 & 0.5 & 0.3500 & 0.1875 & 0.1550 & 0.1588 & 0.2500 & 0.1875 & 0.1675 \\
1 & 0.6 & 0.3725 & 0.1950 & 0.1475 & 0.1588 & 0.2700 & 0.1988 & 0.2012 \\
1 & 0.7 & 0.3575 & 0.1950 & 0.1276 & 0.1012 & 0.3125 & 0.2950 & 0.2879 \\
1 & 0.8 & 0.3625 & 0.2550 & 0.0861 & 0.1046 & 0.3137 & 0.2063 & 0.2975 \\
\midrule
2 & 0.2 & 0.2750 & 0.1222 & 0.1750 & 0.1361 & 0.1800 & 0.0400 & 0.0750 \\
2 & 0.3 & 0.3325 & 0.1750 & 0.1625 & 0.1500 & 0.1912 & 0.0700 & 0.0450 \\
2 & 0.4 & 0.3325 & 0.1650 & 0.1613 & 0.1375 & 0.2137 & 0.1325 & 0.1600 \\
2 & 0.5 & 0.3962 & 0.1567 & 0.1425 & 0.1350 & 0.2537 & 0.1825 & 0.1779 \\
2 & 0.6 & 0.3250 & 0.2125 & 0.1000 & 0.1512 & 0.2767 & 0.2612 & 0.2300 \\
2 & 0.7 & 0.3237 & 0.1450 & 0.1197 & 0.1137 & 0.2937 & 0.2075 & 0.2388 \\
2 & 0.8 & 0.3292 & 0.1963 & 0.1162 & 0.0971 & 0.2854 & 0.2025 & 0.2037 \\
\midrule
3 & 0.2 & 0.2800 & 0.1300 & 0.1400 & 0.1267 & 0.1700 & 0.0500 & 0.0900 \\
3 & 0.3 & 0.3000 & 0.1350 & 0.1700 & 0.1375 & 0.1975 & 0.0850 & 0.0588 \\
3 & 0.4 & 0.3342 & 0.1605 & 0.1387 & 0.1487 & 0.1525 & 0.0625 & 0.1150 \\
3 & 0.5 & 0.3850 & 0.1550 & 0.1350 & 0.1488 & 0.2025 & 0.0825 & 0.1213 \\
3 & 0.6 & 0.3225 & 0.1713 & 0.1150 & 0.1213 & 0.2238 & 0.2275 & 0.1300 \\
3 & 0.7 & 0.3250 & 0.1675 & 0.0921 & 0.0875 & 0.2450 & 0.1400 & 0.1850 \\
3 & 0.8 & 0.3050 & 0.1500 & 0.0763 & 0.0450 & 0.1450 & 0.2200 & 0.2025 \\
\bottomrule
\end{tabular}
\end{table}

\begin{table}
\centering
\rowcolors{2}{gray!10}{white}
\caption{Single-edit FID, $\H\to \G$ --- mean $\aidbar$ by $(n_{\mathrm{bi}},p_{\mathrm{dir}})$ and edit type.} \label{tab:2}
\begin{tabular}{cc|ccccccc}
\toprule
\textbf{bi} & \textbf{dir} & \textbf{reverse} & \textbf{di to bi} & \textbf{add dir} & \textbf{del dir} & \textbf{bi to di} & \textbf{add bi} & \textbf{del bi} \\ \midrule
1 & 0.2 & 0.3278 & 0.1556 & 0.1692 & 0.1722 & 0.1808 & 0.0875 & 0.0375 \\
1 & 0.3 & 0.3382 & 0.1868 & 0.1750 & 0.1816 & 0.2150 & 0.1475 & 0.0837 \\
1 & 0.4 & 0.3550 & 0.1975 & 0.1650 & 0.1425 & 0.2475 & 0.1050 & 0.1213 \\
1 & 0.5 & 0.3500 & 0.1875 & 0.1573 & 0.1575 & 0.2500 & 0.1875 & 0.1675 \\
1 & 0.6 & 0.3725 & 0.1963 & 0.1500 & 0.1550 & 0.2650 & 0.2037 & 0.2025 \\
1 & 0.7 & 0.3600 & 0.1950 & 0.1360 & 0.1008 & 0.3125 & 0.2950 & 0.2875 \\
1 & 0.8 & 0.3612 & 0.2550 & 0.0917 & 0.0975 & 0.3100 & 0.2112 & 0.2925 \\
\midrule
2 & 0.2 & 0.2750 & 0.1222 & 0.1750 & 0.1361 & 0.1800 & 0.0400 & 0.0750 \\
2 & 0.3 & 0.3325 & 0.1725 & 0.1625 & 0.1500 & 0.1900 & 0.0700 & 0.0400 \\
2 & 0.4 & 0.3325 & 0.1650 & 0.1637 & 0.1387 & 0.2125 & 0.1325 & 0.1600 \\
2 & 0.5 & 0.3937 & 0.1525 & 0.1425 & 0.1325 & 0.2500 & 0.1838 & 0.1762 \\
2 & 0.6 & 0.3250 & 0.2112 & 0.1000 & 0.1475 & 0.2712 & 0.2625 & 0.2250 \\
2 & 0.7 & 0.3225 & 0.1450 & 0.1184 & 0.1100 & 0.2925 & 0.2125 & 0.2350 \\
2 & 0.8 & 0.3250 & 0.1900 & 0.1145 & 0.0925 & 0.2800 & 0.2050 & 0.1975 \\
\midrule
3 & 0.2 & 0.2800 & 0.1300 & 0.1400 & 0.1267 & 0.1700 & 0.0500 & 0.0900 \\
3 & 0.3 & 0.3000 & 0.1350 & 0.1700 & 0.1375 & 0.1975 & 0.0875 & 0.0588 \\
3 & 0.4 & 0.3368 & 0.1605 & 0.1375 & 0.1474 & 0.1525 & 0.0625 & 0.1150 \\
3 & 0.5 & 0.3850 & 0.1525 & 0.1350 & 0.1475 & 0.2025 & 0.0825 & 0.1200 \\
3 & 0.6 & 0.3225 & 0.1700 & 0.1150 & 0.1200 & 0.2225 & 0.2275 & 0.1275 \\
3 & 0.7 & 0.3263 & 0.1675 & 0.0921 & 0.0875 & 0.2450 & 0.1400 & 0.1850 \\
3 & 0.8 & 0.3050 & 0.1500 & 0.0789 & 0.0450 & 0.1400 & 0.2200 & 0.2025 \\
\bottomrule
\end{tabular}
\end{table}

\begin{table}
\centering
\rowcolors{2}{gray!10}{white}
\caption{FID vs.\ edit count, $\G\to \H$ --- mean $\aidbar$ by $(n_{\mathrm{bi}},p_{\mathrm{dir}})$ and number of edits $k$.} \label{tab:3}
\begin{tabular}{cc|ccccc}
\toprule
\textbf{bi} & \textbf{dir} & \textbf{1 edit} & \textbf{2 edits} & \textbf{3 edits} & \textbf{4 edits} & \textbf{5 edits}  \\ \midrule
1 & 0.2 & 0.1611 & 0.2358 & 0.2750 & 0.2925 & 0.3550  \\
1 & 0.3 & 0.1905 & 0.2650 & 0.3575 & 0.3675 & 0.3525  \\
1 & 0.4 & 0.0904 & 0.2575 & 0.4067 & 0.4025 & 0.4850  \\
1 & 0.5 & 0.2081 & 0.3204 & 0.4000 & 0.4800 & 0.4425  \\
1 & 0.6 & 0.2206 & 0.2649 & 0.4379 & 0.5342 & 0.5963  \\
1 & 0.7 & 0.2396 & 0.3675 & 0.4075 & 0.4975 & 0.5912  \\
1 & 0.8 & 0.2323 & 0.2712 & 0.4862 & 0.5225 & 0.5775  \\
\midrule
2 & 0.2 & 0.1434 & 0.2225 & 0.2550 & 0.3725 & 0.3475  \\
2 & 0.3 & 0.1609 & 0.1800 & 0.2925 & 0.4000 & 0.3925  \\
2 & 0.4 & 0.1861 & 0.2125 & 0.2950 & 0.3850 & 0.4125  \\
2 & 0.5 & 0.2064 & 0.2275 & 0.3725 & 0.4425 & 0.4963  \\
2 & 0.6 & 0.2224 & 0.3100 & 0.3650 & 0.4450 & 0.5150  \\
2 & 0.7 & 0.2061 & 0.2538 & 0.4075 & 0.5125 & 0.5300  \\
2 & 0.8 & 0.2044 & 0.3662 & 0.3875 & 0.4700 & 0.4838  \\
\midrule
3 & 0.2 & 0.1410 & 0.1675 & 0.2750 & 0.2575 & 0.3550  \\
3 & 0.3 & 0.1549 & 0.2225 & 0.2550 & 0.3925 & 0.3250  \\
3 & 0.4 & 0.1589 & 0.2125 & 0.2475 & 0.3225 & 0.3425  \\
3 & 0.5 & 0.1758 & 0.2825 & 0.3800 & 0.3875 & 0.4425  \\
3 & 0.6 & 0.1874 & 0.2600 & 0.4000 & 0.4088 & 0.4850  \\
3 & 0.7 & 0.1775 & 0.2425 & 0.3850 & 0.4225 & 0.4550  \\
3 & 0.8 & 0.1634 & 0.3075 & 0.3438 & 0.4525 & 0.4925  \\
\bottomrule
\end{tabular}
\end{table}

\begin{table}
\centering
\rowcolors{2}{gray!10}{white}
\caption{FID vs.\ edit count, $\H\to\G$ --- mean $\aidbar$ by $(n_{\mathrm{bi}},p_{\mathrm{dir}})$ and number of edits $k$.} \label{tab:4}
\begin{tabular}{cc|ccccc}
\toprule
\textbf{bi} & \textbf{dir} & \textbf{1 edit} & \textbf{2 edits} & \textbf{3 edits} & \textbf{4 edits} & \textbf{5 edits}  \\ \midrule
1 & 0.2 & 0.1616 & 0.2350 & 0.2750 & 0.2925 & 0.3550  \\
1 & 0.3 & 0.1902 & 0.2662 & 0.3575 & 0.3675 & 0.3525  \\
1 & 0.4 & 0.1906 & 0.2575 & 0.4000 & 0.4025 & 0.4850  \\
1 & 0.5 & 0.2082 & 0.3175 & 0.4000 & 0.4775 & 0.4450  \\
1 & 0.6 & 0.2208 & 0.2621 & 0.4363 & 0.5325 & 0.5963  \\
1 & 0.7 & 0.2410 & 0.3625 & 0.4075 & 0.4975 & 0.5875  \\
1 & 0.8 & 0.2313 & 0.2675 & 0.4862 & 0.5237 & 0.5825  \\
\midrule
2 & 0.2 & 0.1434 & 0.2225 & 0.2550 & 0.3725 & 0.3488  \\
2 & 0.3 & 0.1597 & 0.1850 & 0.2925 & 0.4000 & 0.3925  \\
2 & 0.4 & 0.1865 & 0.2125 & 0.2950 & 0.3850 & 0.4125  \\
2 & 0.5 & 0.2045 & 0.2300 & 0.3725 & 0.4425 & 0.4925  \\
2 & 0.6 & 0.2207 & 0.3113 & 0.3650 & 0.4375 & 0.5150  \\
2 & 0.7 & 0.2052 & 0.2525 & 0.4050 & 0.5138 & 0.5312  \\
2 & 0.8 & 0.2007 & 0.3650 & 0.3875 & 0.4700 & 0.4825  \\
\midrule
3 & 0.2 & 0.1410 & 0.1675 & 0.2750 & 0.2575 & 0.3550  \\
3 & 0.3 & 0.1552 & 0.2250 & 0.2550 & 0.3925 & 0.3250  \\
3 & 0.4 & 0.1589 & 0.2137 & 0.2475 & 0.3225 & 0.3425  \\
3 & 0.5 & 0.1750 & 0.2825 & 0.3800 & 0.3875 & 0.4425  \\
3 & 0.6 & 0.1865 & 0.2600 & 0.4000 & 0.4075 & 0.4850  \\
3 & 0.7 & 0.1777 & 0.2425 & 0.3850 & 0.4225 & 0.4525  \\
3 & 0.8 & 0.1631 & 0.3075 & 0.3425 & 0.4525 & 0.4925  \\
\bottomrule
\end{tabular}
\end{table}

\begin{table}
    \centering
    \caption{Pooled single-edit impact --- mean $\aidbar$ (± s.e.) by edit type across all settings.}\label{tab:breakdown}
    \begin{tabular}{c|ccccccc}
    \toprule
        & \textbf{reverse dir} & \textbf{dir to bi} & \textbf{add dir} & \textbf{del dir} & \textbf{bi to dir} & \textbf{add bi} & \textbf{del bi} \\ \midrule
      \textbf{mean} & 0.3357 & 0.1726 & 0.1376 & 0.1303 & 0.2288 & 0.1526 & 0.1531\\
      \textbf{std.\ error} & 0.0048 & 0.0049 & 0.0039 & 0.0037 & 0.0060 & 0.0080 & 0.0073 \\
      \bottomrule
    \end{tabular}
\end{table}

\section{Software}
We use Python \cite{vanrossum2009python} and list all used packages in \cref{tab:software}. Most of them are open software, with two exceptions: \texttt{dct-policy} and \texttt{sep\_distances}, which are publicly available on \texttt{github.com}, but which do not include an explicit license.  

\begin{table}
\centering
\rowcolors{2}{gray!10}{white}
\caption{Overview of software and packages used in this work.}\label{tab:software}
\smallskip
\begin{tabular}{l l l}
  \toprule
    \textbf{Name} & \textbf{Reference} & \textbf{License}  \\ 
    \midrule
    Python          &  \cite{vanrossum2009python}
                    & \href{https://docs.python.org/3/license.html#psf-license}{PSF License} \\
    Numpy           &  \cite{harris2020numpy}
                    & \href{https://github.com/numpy/numpy/blob/main/LICENSE.txt}{BSD-style license} \\
    Matplotlib      &  \cite{hunter2007matplotlib}
                    & \href{https://matplotlib.org/stable/users/project/license.html}{modified PSF} \\
    Scikit-learn    &     \cite{pedregosa2011scikit}
                    & \href{https://github.com/scikit-learn/scikit-learn/blob/main/COPYING}{BSD 3-Clause} \\
    SLURM           & \cite{yoo2003slurm}
                    & \href{https://github.com/SchedMD/slurm/tree/master?tab=License-1-ov-file}{modified GNU GPL v2} \\
    networkx        & \cite{networkx}
                    & \href{https://raw.githubusercontent.com/networkx/networkx/master/LICENSE.txt}{BSD 3-Clause} \\
    
    sep\_distances  & \cite{wahl2025separationbaseddistancemeasurescausal}
                    & \href{https://raw.githubusercontent.com/networkx/networkx/master/LICENSE.txt}{no license} \\

    Ananke  & \cite{lee2023anankepythonpackagecausal}
                    & \href{https://gitlab.com/causal/ananke/}{Apache 2.0} \\

    Pandas          &  \cite{reback2020pandas}
                    & \href{https://github.com/pandas-dev/pandas/blob/main/LICENSE}{BSD-style license} \\
    dct-policy &  \cite{squires2020activestructurelearningcausal}
                    & \href{https://github.com/csquires/dct-policy}{no license} \\
    SciPy           &  \cite{virtanen2020scipy}
                    & \href{https://github.com/scipy/scipy/blob/main/LICENSE.txt}{BSD 3-Clause} \\
    \bottomrule
\end{tabular}
\end{table}
\end{document}